\documentclass[runningheads]{llncs}
\usepackage{graphicx}
\usepackage{amsmath,amssymb} 
\usepackage{color}
\usepackage[width=122mm,left=12mm,paperwidth=146mm,height=193mm,top=12mm,paperheight=217mm]{geometry}

\usepackage[square,numbers,sort&compress]{natbib}

\usepackage[bookmarks=false,pagebackref=true,breaklinks=true,citecolor=blue,colorlinks=true]{hyperref}
\usepackage{multirow}
\usepackage{dsfont}
\usepackage{booktabs}
\usepackage{caption}
\usepackage{array}
\usepackage{colortbl}
\usepackage[normalem]{ulem}

\def\ie{\emph{i.e.}}
\def\etal{\emph{et al.}}
\def\etc{\emph{etc.}}
\def\eg{\emph{eg.}}
\definecolor{Gray}{gray}{0.7}

\begin{document}
\pagestyle{headings}
\makeatother
\title{Shuffle and Learn: Unsupervised Learning using Temporal Order Verification}

\titlerunning{Shuffle and Learn: Unsupervised Learning using Temporal Order Verification}

\authorrunning{Ishan Misra, C. Lawrence Zitnick and Martial Hebert}

\author{Ishan Misra$^{1}$ \quad \quad C. Lawrence Zitnick$^{2}$ \quad \quad Martial Hebert$^{1}$}
\institute{The Robotics Institute, Carnegie Mellon University
\and 
Facebook AI Research\\
   \email{ \{imisra, hebert\}@cs.cmu.edu, zitnick@fb.com}
   }

\maketitle

\begin{abstract}
In this paper, we present an approach for learning a visual representation from the raw spatiotemporal signals in videos. Our representation is learned without supervision from semantic labels. We formulate our method as an unsupervised sequential verification task, i.e., we determine whether a sequence of frames from a video is in the correct temporal order. With this simple task and no semantic labels, we learn a powerful visual representation using a Convolutional Neural Network (CNN). The representation contains complementary information to that learned from supervised image datasets like ImageNet.
Qualitative results show that our method captures information that is temporally varying, such as human pose.
When used as pre-training for action recognition, our method gives significant gains over learning without external data on benchmark datasets like UCF101 and HMDB51. To demonstrate its sensitivity to human pose, we show results for pose estimation on the FLIC and MPII datasets that are competitive, or better than approaches using significantly more supervision. Our method can be combined with supervised representations to provide an additional boost in accuracy.
\end{abstract}

\section{Introduction}
Sequential data provides an abundant source of information in the form of auditory and visual percepts. Learning from the observation of sequential data is a natural and implicit process for humans~\cite{cleeremans1991learning,reber1989implicit,cleeremans1993mechanisms}. It informs both low level cognitive tasks and high level abilities like decision making and problem solving~\cite{sun2001implicit}. For instance, answering the question ``Where would the moving ball go?'', requires the development of basic cognitive abilities like prediction from sequential data like video~\cite{baker2014learning}.

In this paper, we explore the power of spatiotemporal signals, \ie, videos, in the context of computer vision. To study the information available in a video signal in isolation, we ask the question: How does an agent learn from the spatiotemporal structure present in video without using supervised semantic labels? Are the representations learned using the unsupervised spatiotemporal information present in videos meaningful? And finally, are these representations complementary to those learned from strongly supervised image data? In this paper, we explore such questions by using a sequential learning approach.

Sequential learning is used in a variety of areas such as speech recognition, robotic path planning, adaptive control algorithms, \etc \ These approaches can be broadly categorized~\cite{sun2001sequence} into two classes: prediction and verification. In sequential prediction, the goal is to predict the signal given an input sequence. A popular application of this in Natural Language Processing (NLP) is `word2vec' by Mikolov \etal~\cite{mikolov2013efficient,mikolov2013distributed} that learns distributional representations~\cite{firth1957synopsis}. Using the continuous bag-of-words (CBOW) task, the model learns to predict a missing word given a sequence of surrounding words. The representation that results from this task has been shown to be semantically meaningful~\cite{mikolov2013efficient}. Unfortunately, extending the same technique to predict video frames is challenging. Unlike words that can be represented using limited-sized vocabularies, the space of possible video frames is extremely large~\cite{doersch-context}, \eg, predicting pixels in a small $256\times256$ image leads to $256^{2\times3\times256}$ hypotheses! To avoid this complex task of predicting high-dimensional video frames, we use sequential verification.

\begin{figure}[!t]
\centering
\includegraphics[width=1.0\textwidth]{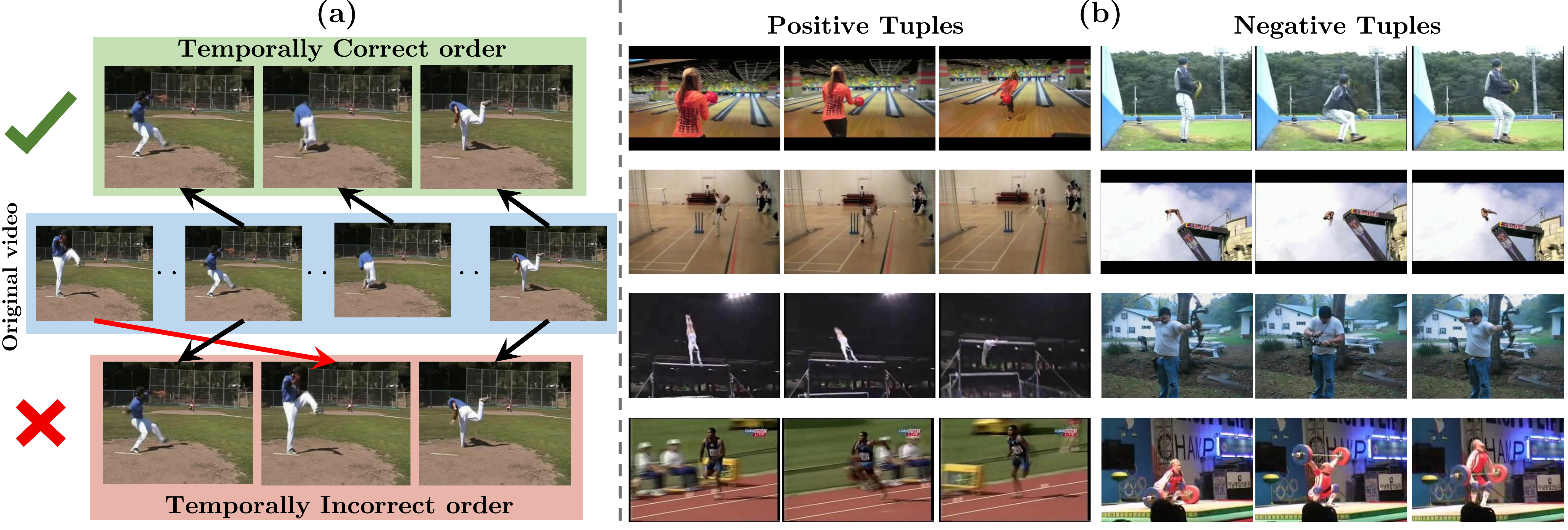}
\caption{\footnotesize{\textbf{(a)} A video imposes a natural temporal structure for visual data. In many cases, one can easily verify whether frames are in the correct temporal order (shuffled or not). Such a simple sequential verification task captures important spatiotemporal signals in videos. We use this task for unsupervised pre-training of a Convolutional Neural Network (CNN). \textbf{(b)} Some examples of the automatically extracted positive and negative tuples used to formulate a classification task for a CNN.}}
\label{fig:teaser}
\end{figure}

In sequential verification, one predicts the `validity' of the sequence, rather than individual items in the sequence. In this paper, we explore the task of determining whether a given sequence is `temporally valid', \ie, whether a sequence of video frames are in the correct temporal order, Figure \ref{fig:teaser}. We demonstrate that this binary classification problem is capable of learning useful visual representations from videos. Specifically, we explore their use in the well understood tasks of human action recognition and pose estimation. But why are these simple sequential verification tasks useful for learning? Determining the validity of a sequence requires reasoning about object transformations and relative locations through time. This in turn forces the representation to capture object appearances and deformations.

We use a Convolutional Neural Network (CNN)~\cite{lecun1989backpropagation} for our underlying feature representation. The CNN is applied to each frame in the sequence and trained ``end-to-end'' from random initialization. The sequence verification task encourages the CNN features to be both visually and temporally grounded. We demonstrate the effectiveness of our unsupervised method on benchmark action recognition datasets UCF101~\cite{ucf101} and HMDB51~\cite{hmdb51}, and the FLIC~\cite{modec13} and MPII~\cite{andriluka14cvpr} pose estimation datasets. Using our simple unsupervised learning approach for pre-training, we show a significant boost in accuracy over learning CNNs from scratch with random initialization. In fact, our unsupervised approach even outperforms pre-training with some supervised training datasets. In action recognition, improved performance can be found by combining existing supervised image-based representations with our unsupervised representation. By training on action videos with humans, our approach learns a representation sensitive to human pose. Remarkably, when applied to pose estimation, our representation is competitive with pre-training on significantly larger supervised training datasets~\cite{ImageNet}.

\section{Related Work}

Our work uses unlabeled video sequences for learning representations. Since this source of supervision is `free', our work can be viewed as a form of unsupervised learning. Unsupervised representation learning from single images is a popular area of research in computer vision. A significant body of unsupervised learning literature uses hand-crafted features and clustering based approaches to discover objects~\cite{faktor2012clustering,sivic2005discovering,russell2006using}, or mid-level elements~\cite{singh2012unsupervised,juneja2013blocks,doersch2013mid,li2013harvesting,sun2013learning}. Deep learning methods like auto-encoders~\cite{olshausen,bengio2013deep,vincent2008extracting}, Deep Boltzmann Machines~\cite{salakhutdinov2009deep}, variational methods~\cite{kingma2013auto,rezende2014stochastic}, stacked auto-encoders~\cite{lee2006efficient,bengio2007greedy}, and others~\cite{le2013building,ssgan16} learn representations directly from images. These methods learn a representation by estimating latent parameters that help reconstruct the data, and may regularize the learning process by priors such as sparsity~\cite{olshausen}. Techniques in~\cite{sermanet2013pedestrian,doersch-context} scale unsupervised learning to large image datasets showing its usefulness for tasks such as pedestrian detection~\cite{sermanet2013pedestrian} and object detection~\cite{doersch-context}. In terms of using `context' for learning, our work is most similar to~\cite{doersch-context} which uses the spatial context in images. While these approaches are unsupervised, they do not use videos and cannot exploit the temporal structure in them.
\begin{figure}[!t]
\centering
\includegraphics[width=0.95\textwidth]{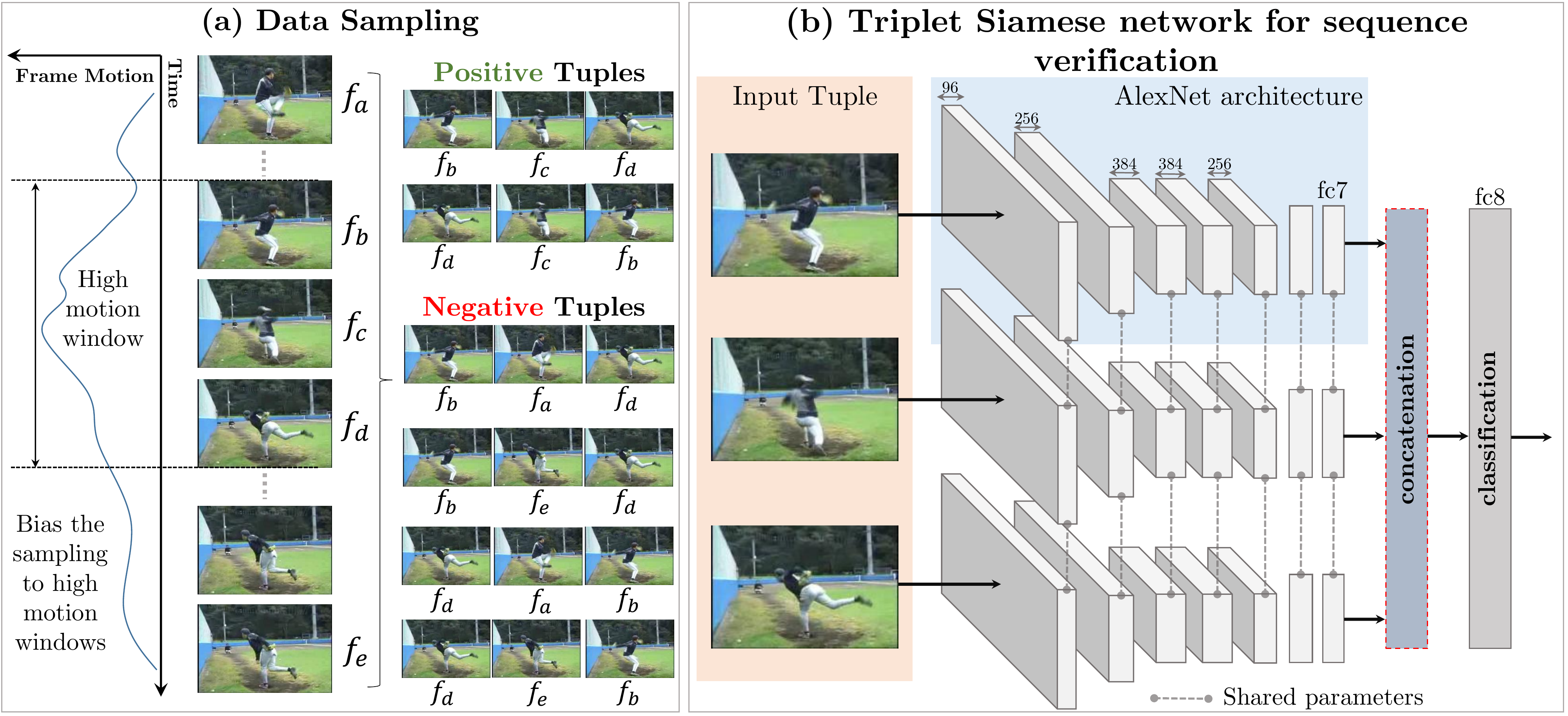}
\caption{\footnotesize{\textbf{(a)} We sample tuples of frames from high motion windows in a video. We form positive and negative tuples based on whether the three input frames are in the correct temporal order. \textbf{(b)} Our triplet Siamese network architecture has three parallel network stacks with shared weights upto the \texttt{fc7} layer. Each stack takes a frame as input, and produces a representation at the \texttt{fc7} layer. The  concatenated \texttt{fc7} representations are used to predict whether the input tuple is in the correct temporal order.}}
\label{fig:narch}
\end{figure}
Our work is most related to work in unsupervised learning from videos~\cite{dinesh-equiv,dinesh-slow,mobahi2009deep,isola2015learning,hadsell2006dimensionality}. Traditional methods in this domain utilize the spatiotemporal continuity as regularization for the learning process. Since visual appearance changes smoothly in videos, a common constraint is enforcing temporal smoothness of features~\cite{foldiak1991learning,wiskott2002slow,mobahi2009deep,goroshin2015unsupervised,hadsell2006dimensionality}. Zhang \etal~\cite{sfa-action}, in particular, show how such constraints are useful for action recognition. Moving beyond just temporal smoothness,~\cite{dinesh-slow} enforces additional `steadiness' constraints on the features so that the change of features across frames is meaningful. Our work, in contrast, does not explicitly impose any regularizations on the features. Other reconstruction-based learning approaches include that of Goroshin \etal~\cite{goroshin2015unsupervised} who use a generative model to predict video frames and Srivastava \etal~\cite{srivastava-lstms} who use LSTMs~\cite{lstm}. Unlike our method, these works~\cite{taylor2010convolutional,goroshin2015unsupervised,srivastava-lstms,mobahi2009deep} explicitly predict individual frames, but do not explore large image sizes or datasets.~\cite{pred-berg,vondrick2015anticipating} also consider the task of predicting the future from videos, but consider it as their end task and do not use it for unsupervised pre-training.

Several recent papers~\cite{dinesh-equiv,pulkit-ego,pred-berg} use egomotion constraints from video to further constrain the learning. Jayaraman \etal~\cite{dinesh-equiv} show how they can learn equivariant transforms from such constraints. Similar to our work, they use full video frames for learning with little pre-processing. Owens \etal~\cite{owens2016visually} use audio signals from videos to learn visual representations. Another line of work~\cite{wang-gupta} uses video data to mine patches which belong to the same object to learn representations useful for distinguishing objects. Typically, these approaches require significant pre-processing to create this task. While our work also uses videos, we explore them in the spirit of sequence verification for action recognition which learns from the raw video with very little pre-processing.

We demonstrate the effectiveness of our unsupervised pre-training using two extensively studied vision tasks - action recognition and pose estimation. These tasks have well established benchmark datasets~\cite{ucf101,hmdb51,modec13,andriluka14cvpr}. As it is beyond the scope of this paper, we refer the reader to~\cite{poppe2010survey} for a survey on action recognition, and~\cite{perez2014survey} for a survey on pose estimation.

\section{Our Approach}

Our goal is to learn a feature representation using only the raw spatiotemporal signal naturally available in videos. We learn this representation using a sequential verification task and focus on videos with human actions. Specifically, as shown in Figure~\ref{fig:teaser}, we extract a tuple of frames from a video, and ask whether the frames are in the correct temporal order. In this section, we begin by motivating our use of sequential tasks and how they use the temporal structure of videos. We then describe how positive and negative tuples are sampled from videos, and describe our model.

\subsection{Task motivation}
\label{sec:intuition}
When using only raw videos as input, sequential verification tasks offer a promising approach to unsupervised learning. In addition to our approach described below, several alternative tasks are explored in Section \ref{sec:twoframes}. The goal of these tasks is to encourage the model to reason about the motion and appearance of the objects, and thus learn the temporal structure of videos. Example tasks may include reasoning about the ordering of frames, or determining the relative temporal proximity of frames. For tasks that ask for the verification of temporal order, how many frames are needed to determine a correct answer? If we want to determine the correct order from just two frames, the question may be ambiguous in cases where cyclical motion is present. For example, consider a short video sequence of a person picking up a coffee cup. Given two frames the temporal order is ambiguous; the person may be picking the coffee cup up, or placing it down.

To reduce such ambiguity, we propose sampling a three frame tuple, and ask whether the tuple's frames are correctly ordered. While theoretically, three frames are not sufficient to resolve cyclical ambiguity~\cite{shannon1949communication}, we found that combining this with smart sampling (Section~\ref{sec:sampling-overview}) removes a significant portion of ambiguous cases. We now formalize this problem into a classification task. Consider the set of frames $\{f_1, \ldots, f_n \}$ from an unlabeled video $\mathcal{V}$. We consider the tuple $(f_b,f_c,f_d)$ to be in the correct temporal order (class 1, positive tuple) if the frames obey either ordering $b < c < d$ or $d < c < b$, to account for the directional ambiguity in video clips. Otherwise, if $b < d < c$ or $c < b < d$,  we say that the frames are not in the correct temporal order (class 0, negative tuple).

\subsection{Tuple sampling}
\label{sec:sampling-overview}

A critical challenge when training a network on the three-tuple ordering task is how to sample positive and negative training instances. A naive method may sample the tuples uniformly from a video. However, in temporal windows with very little motion it is hard to distinguish between a positive and a negative tuple, resulting in many ambiguous training examples. Instead, we only sample tuples from temporal windows with high motion. As Figure~\ref{fig:narch} shows, we use coarse frame level optical flow \cite{farneback2003two} as a proxy to measure the motion between frames. We treat the average flow magnitude per-frame as a weight for that frame, and use it to bias our sampling towards high motion windows. This ensures that the classification of the tuples is not ambiguous. Figure~\ref{fig:teaser} (b) shows examples of such tuples. 

To create positive and negative tuples, we sample five frames $(f_a, f_b, f_c, f_d, f_e)$ from a temporal window such that $a < b < c < d < e$ (see Figure~\ref{fig:narch} (a)). Positive instances are created using $(f_b, f_c, f_d)$, while negative instances are created using $(f_b, f_a, f_d)$ and $(f_b, f_e, f_d)$. Additional training examples are also created by inverting the order of all training instances, \eg, $(f_d,f_c,f_b)$ is positive. During training it is critical to use the same beginning frame $f_b$ and ending frame $f_d$ while only changing the middle frame for both positive and negative examples. Since only the middle frame changes between training examples, the network is encouraged to focus on this signal to learn the subtle difference between positives and negatives, rather than irrelevant features.

To avoid sampling ambiguous negative frames $f_a$ and $f_e$, we enforce that the appearance of the positive $f_c$ frame is not too similar (measured by SSD on RGB pixel values) to $f_a$ or $f_e$. These simple conditions eliminated most ambiguous examples. We provide further analysis of sampling data in Section~\ref{sec:sampling-ablation}.

\subsection{Model Parametrization and Learning}
\label{sec:narch}
To learn a feature representation from the tuple ordering task, we use a simple triplet Siamese network. This network has three parallel stacks of layers with shared parameters (Figure~\ref{fig:narch}). Every network stack follows the standard CaffeNet \cite{caffe} (a slight modification of AlexNet \cite{alexnet}) architecture from the \texttt{conv1} to the \texttt{fc7} layer. Each stack takes as input one of the frames from the tuple and produces a representation at the \texttt{fc7} layer. The three \texttt{fc7} outputs are concatenated as input to a linear classification layer. The classification layer can reason about all three frames at once and predict whether they are in order or not (two class classification). Since the layers from \texttt{conv1} to \texttt{fc7} are shared across the network stacks, the Siamese architecture has the same number of parameters as AlexNet barring the final \texttt{fc8} layer. We update the parameters of the network by minimizing the regularized cross-entropy loss of the predictions on each tuple. While this network takes three inputs at training time, during testing we can obtain the \texttt{conv1} to \texttt{fc7} representations of a single input frame by using just one stack, as the parameters across the three stacks are shared.

\section{Empirical ablation analysis}
\label{sec:ablation}
In this section (and in the Appendix), we present experiments to analyze the various design decisions for training our network. 
In Sections \ref{sec:main-exp} and \ref{sec:pose}, we provide results on both action recognition and pose estimation. 

\par \noindent \textbf{Dataset:} We report all our results using split 1 of the benchmark UCF101~\cite{ucf101} dataset. This dataset contains videos for 101 action categories with $\sim9.5$k videos for training and $\sim3.5$k videos for testing. Each video has an associated action category label. The standard performance metric for action recognition on this dataset is classification accuracy.

\par \noindent \textbf{Details for unsupervised pre-training:} For unsupervised pre-training, we do not use the semantic action labels. We sample about $900$k tuples from the UCF101 training videos. We randomly initialize our network, and train for $100$k iterations with a fixed learning rate of $10^{-3}$ and mini-batch size of $128$ tuples. Each tuple consists of 3 frames. Using more (4, 5) frames per tuple did not show significant improvement. We use batch normalization~\cite{batchnorm}.

\par \noindent \textbf{Details for Action Recognition:}
The spatial network from~\cite{simonyan-2stream} is a well-established method of action recognition that uses only RGB appearance information. The parameters of the spatial network are initialized with our unsupervised pre-trained network.
We use the provided action labels per video and follow the training and testing protocol as suggested in\cite{simonyan-2stream,cuhk-2stream}. Briefly, for training we form mini-batches by sampling random frames from videos. At test time, $25$ frames are uniformly sampled from each video. Each frame is used to generate $10$ inputs after fixed cropping and flipping (5 crops $\times$ 2 flips), and the prediction for the video is an average of the predictions across these $25 \times 10$ inputs. 
We use the CaffeNet architecture for its speed and efficiency. We initialize the network parameters up to the \texttt{fc7} layer using the parameters from the unsupervised pre-trained network, and initialize a new \texttt{fc8} layer for the action recognition task. We finetune the network following~\cite{simonyan-2stream} for $20$k iterations with a batch size of $256$, and learning rate of $10^{-2}$ decaying by $10$ after $14$k iterations, using SGD with momentum of $0.9$, and dropout of $0.5$. While~\cite{simonyan-2stream} used the wider VGG-M-2048~\cite{chatfield2014return} architecture, we found that their parameters transfer to CaffeNet because of the similarities in their architectures.

\begin{table}[!t]
\centering
\caption{\footnotesize{We study the effect of our design choices such as temporal sampling parameters, and varying class ratios for unsupervised pre-training. We measure the tuple prediction accuracy on a held out set from UCF101. We also show action classification results after finetuning the models on the UCF101 action recognition task (split 1).}}
\label{tbl:ablation}
\parbox{.45\linewidth}{
\centering
\caption*{\footnotesize{(a) Varying temporal sampling}}
\label{tbl:temporal-sampling}
\begin{tabular}[b]{@{}cccccc@{}}
\toprule
$\tau_{\max}$ & $\tau_{\min}$ && Tuple Pred. && Action Recog. \\
\midrule
30 & 15 && 60.2 && 47.2 \\
60 & 15 && \textbf{72.1} &&  \textbf{50.9} \\
60 & 60 && 64.3  && 49.1 \\
\bottomrule
\end{tabular}
} \hfill
\parbox{.45\linewidth}{
\centering
\caption*{\footnotesize{(b) Varying class ratios}}
\label{tbl:class-ratios}
\begin{tabular}[b]{@{}cccccc@{}}
\toprule
\multicolumn{2}{c}{Class Ratio} && Tuple Pred. && Action Recog.\\
Neg& Pos &&  &  \\
\midrule
0.5& 0.5 && 52.1 && 38.1 \\
0.65& 0.35 && 68.5 && 45.5 \\
0.75& 0.25 && \textbf{72.1} && \textbf{50.9} \\
0.85& 0.15 && 67.7 && 48.6 \\
\bottomrule
\end{tabular}
}
\end{table}

\subsection{Sampling of data}
\label{sec:sampling-ablation}
In this section we study the impact of sampling parameters described in Section~\ref{sec:sampling-overview} on the unsupervised pre-training task. 
 We denote the maximum distance between frames of positive tuples by $\tau_{\max} = |b-d|$. This parameter controls the `difficulty' of positives: a very high value makes it difficult to see correspondence across the positive tuple, and a very low value gives almost identical frames and thus very easy positives. Similarly, we compute the minimum distance between the frames $f_a$ and $f_e$ used for negative tuples to the other frames by $\tau_{\min} = \min(|a-b|,|d-e|)$.  This parameter controls the difficulty of negatives with a low value making them harder, and a high value making them easier.

We compute the training and testing accuracy of these networks on the tuple prediction task on held out videos. This held out set is a union of samples using all the temporal sampling parameters. We show results in Table~\ref{tbl:temporal-sampling} (a). We also use these networks for finetuning on the UCF101 action recognition task. Our results show that the tuple prediction accuracy and the performance on the action recognition task are correlated. A large temporal window for positive sampling improves over a smaller temporal window (Rows 1 and 2), while a large window for negative sampling hurts performance (Rows 2 and 3).

\begin{figure}[!t]
\centering
\includegraphics[width=0.9\textwidth]{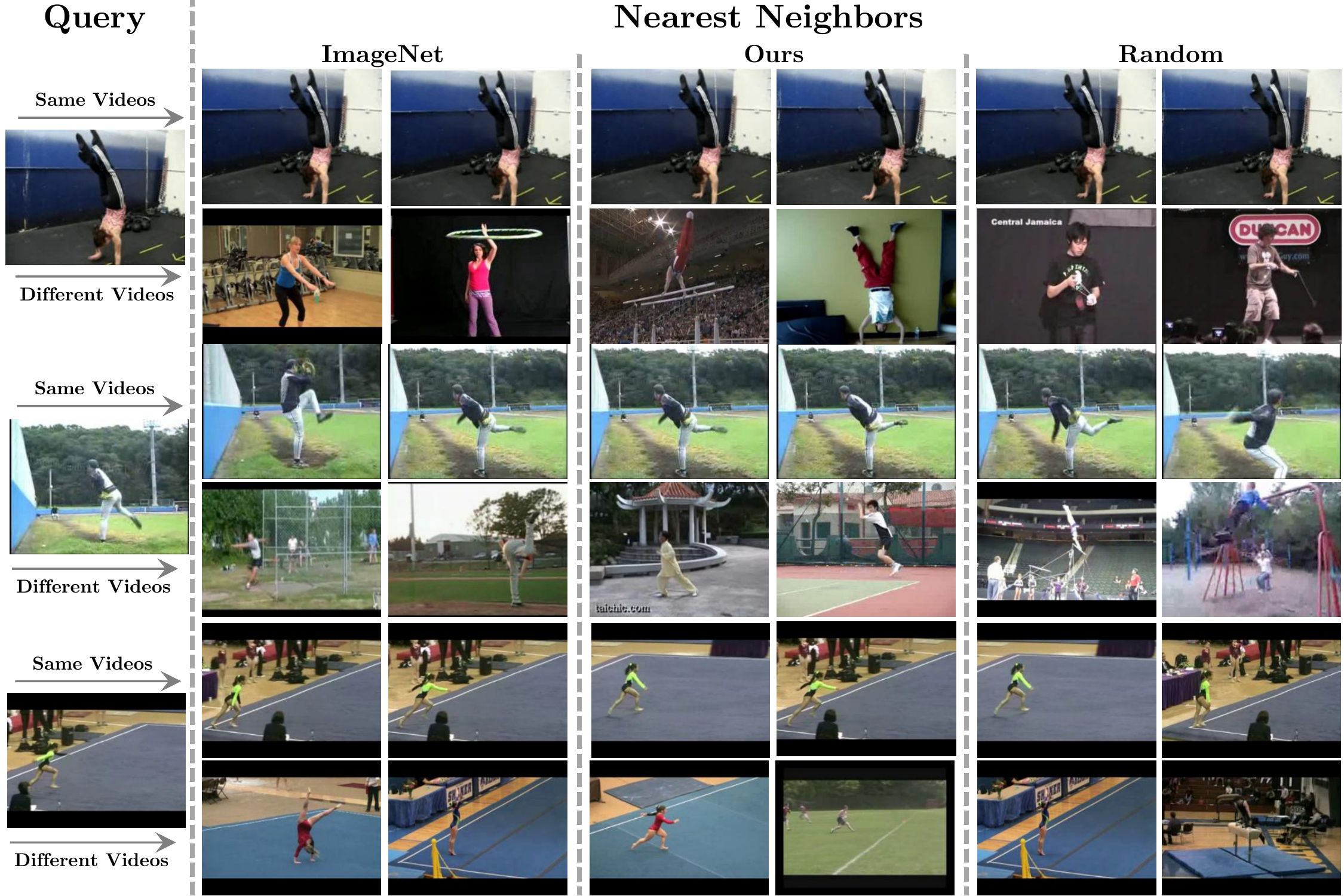}
\caption{\footnotesize{We compute nearest neighbors using \texttt{fc7} features on the UCF101 dataset. We compare these results across three networks: pre-trained on ImageNet, pre-trained on our unsupervised task and a randomly initialized network. We choose a input query frame from a clip and retrieve results from other clips in the dataset. Since the dataset contains multiple clips from the same video we get near duplicate retrievals (first row). We remove these duplicates, and display results in the second row. While ImageNet focuses on the high level semantics, our network captures the human pose.}}
\label{fig:nn-res}
\end{figure}

\subsection{Class ratios in mini-batch}
Another important factor when training the model is the class ratios in each mini-batch. As has been observed empirically~\cite{fast-rcnn,faster-rcnn},  a good class ratio per mini-batch ensures that the model does not overfit to one particular class, and helps the learning process. For these experiments, we choose a single temporal window for sampling and vary only the ratio of positive and negative tuples per mini-batch. We compare the accuracy of these networks on the tuple prediction task on held out videos in Table~\ref{tbl:class-ratios} (b). Additionally, we report the accuracy of these networks after finetuning on the action recognition task. These results show that the class ratio used for unsupervised pre-training can significantly impact learning. It is important to have a larger percentage of negative examples.

\subsection{What does the temporal ordering task capture?}
\label{sec:qual-res}
\subsubsection{Nearest Neighbor retrieval}
We retrieve nearest neighbors using our unsupervised features on the UCF101 dataset and compare them in Figure~\ref{fig:nn-res} to retrievals by the pre-trained ImageNet features, and a randomly initialized network. Additional examples are shown in the supplementary materials. We pick an input query frame from a clip and retrieve neighbors from other clips in the UCF101 dataset. Since the UCF101 dataset has clips from the same video, the first set of retrievals (after removing frames from the same input clip) are near duplicates which are not very informative (notice the random network's results). We remove these near-duplicates by computing the sum of squared distances (SSD) between the frames, and display the top results in the second row of each query. These results make two things clear: 1) the ImageNet pre-trained network focuses on scene semantics 2) Our unsupervised pre-trained network focuses on the pose of the person. This would seem to indicate that the information captured by our unsupervised pre-training is complementary to that of ImageNet. Such behavior is not surprising, if we consider our network was trained without semantic labels, and must reason about spatiotemporal signals for the tuple verification task.

\subsubsection{Visualizing \texttt{pool5} unit responses}
\begin{figure}[!ht]
\centering
\includegraphics[width=0.9\textwidth]{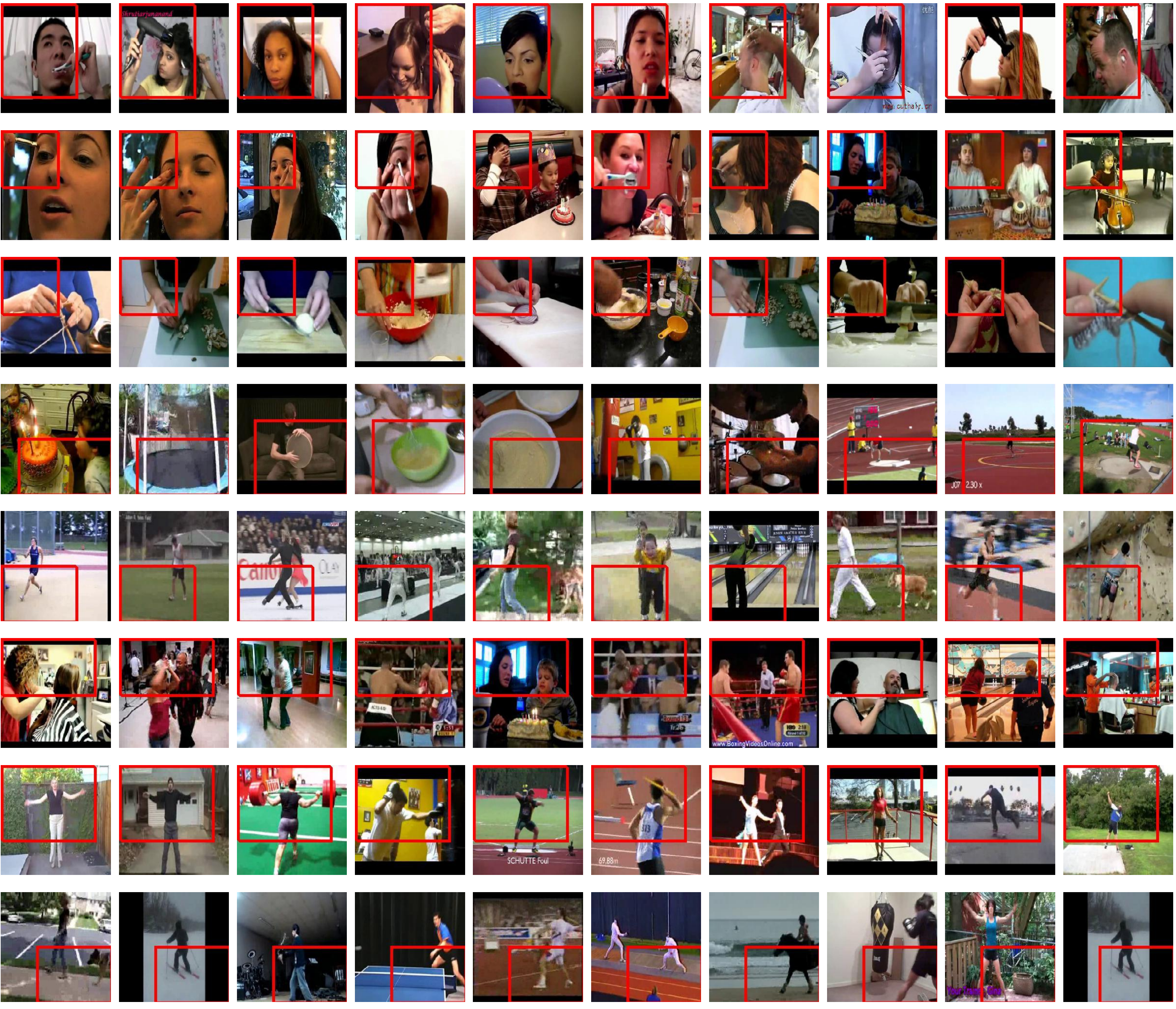}
\caption{In each row we display the top image regions for a unit from the \texttt{pool5} layer. We follow the method in~\cite{girshick2014rich} and display the receptive fields (marked in red boxes) for these units. As our network is trained on human action recognition videos, many units show preference for human body parts and pose.}
\label{fig:pool5viz}
\end{figure}

We analyze the feature representation of the unsupervised network trained using the tuple prediction task on UCF101. Following the procedure of~\cite{girshick2014rich} we show the top regions for \texttt{pool5} units alongwith their receptive field in Figure~\ref{fig:pool5viz}. This gives us insight into the network's internal feature representation and shows that many units show preference for human body parts and pose. This is not surprising given that our network is trained on videos of human action recognition, and must reason about human movements for the tuple ordering task.

\section{Additional Experiments on Action Recognition}
\label{sec:main-exp}

The previous experiments show that the unsupervised task learns a meaningful representation. In this section we compare our unsupervised method against existing baseline methods and present more quantitative results. We organize our experiments as follows: 
1) Comparing our unsupervised method to learning from random initialization. 2) Exploring other unsupervised baselines and comparing our method with them. 3) Combining our unsupervised representation learning method with a supervised image representation. Additional experiments are in the supplementary material. We now describe the common experimental setup.

\par \noindent \textbf{Datasets and Evaluation:} We use the UCF101~\cite{ucf101} dataset which was also used for our ablation analysis in Section~\ref{sec:ablation} and measure accuracy on the 101 action classification task. Additionally, we use the HMDB51~\cite{hmdb51} dataset for action recognition. This dataset contains 3 splits for train/test, each with about $3.4$k videos for train and $1.4$k videos for testing. Each video belongs to one of 51 action categories, and performance is evaluated by measuring classification accuracy. We follow the same train/test protocols for both UCF101 and HMDB51 as described in Section~\ref{sec:ablation}. Note that the UCF101 dataset is about $2.5\times$ larger than the HMDB51 dataset.

\par \noindent \textbf{Implementation details for pre-training:} We use tuples sampled using $\tau_{\max} = 60$ and $\tau_{\min} = 15$ as described in Section~\ref{sec:ablation}. The class ratio of positive examples per mini-batch is $25\%$. The other parameters for training/finetuning are kept unchanged from Section~\ref{sec:ablation}.

\par \noindent \textbf{Action recognition details:} As in Section~\ref{sec:ablation}, we use the CaffeNet architecture and the parameters from~\cite{simonyan-2stream} for both training from scratch and finetuning. We described the finetuning parameters in Section~\ref{sec:ablation}.
For training from random initialization (or `scratch'), we train for $80$k iterations with an initial learning rate of $10^{-2}$, decaying by a factor of $10$ at steps $50$k and $70$k. The other training parameters (momentum, batch size etc.) are kept the same as in finetuning. We use the improved data augmentation scheme (different aspect-ratio, fixed crops) from~\cite{cuhk-2stream} for all our methods and baselines. Note that we train or finetune all the layers of the network for all methods, including ours.

\subsection{Unsupervised pre-training or random initialization?}
\label{sec:unsup-finetune}
In these experiments we study the advantage of unsupervised pre-training for action recognition in comparison to learning without any pre-training. We use our tuple prediction task to train a network starting from random initialization on the train split of UCF101. The unsupervised pre-trained network is finetuned on both the UCF101 and HMDB51 datasets for action recognition and compared against learning from scratch (without pre-training). We report the performance in Table~\ref{tbl:unsup-vs-scratch}. Our unsupervised pre-training shows a dramatic \textbf{improvement of $\mathbf{+12.4\%}$} over training from scratch in UCF101 and a significant gain of $+4.7\%$ in HMDB51. This impressive gain demonstrates the informativeness of the unsupervised tuple verification task. On HMDB51, we additionally finetune a network which was trained from scratch on UCF101 and report its performance in Table~\ref{tbl:unsup-vs-scratch} indicated by `UCF supervised'. We see that this network performs worse than our unsupervised pre-trained network. The UCF101 and HMDB51 have only 23 action classes in common~\cite{simonyan-2stream} and we hypothesize that the poor performance is due to the scratch UCF101 network being unable to generalize to actions from HMDB51. For reference, a model pre-trained on the supervised ImageNet dataset~\cite{ImageNet,ILSVRC15} and finetuned on UCF101 gives $67.1\%$ accuracy, and ImageNet finetuned on HMDB51 gives an accuracy of $28.5\%$.

\begin{table}[!t]
\centering
\caption{\footnotesize{Mean classification accuracies over the 3 splits of UCF101 and HMDB51 datasets. We compare different initializations and finetune them for action recognition.}}
\label{tbl:unsup-vs-scratch}
\begin{tabular}[b]{@{}lcc@{}}
Dataset & Initialization & Mean Accuracy\\
\toprule
UCF101 & Random & 38.6 \\
 & (Ours) Tuple verification & \textbf{50.2}\\
\midrule
HMDB51 & Random & 13.3 \\
 & UCF Supervised & 15.2\\
& (Ours) Tuple verification  & \textbf{18.1}\\
\bottomrule
\end{tabular}
\end{table}

\subsection{Unsupervised Baselines}
\label{sec:twoframes}
In this section, we enumerate a variety of alternative verification tasks that use only video frames and their temporal ordering. For each task, we use a similar frame sampling procedure to the one described in Section~\ref{sec:sampling-ablation}. 
We compare their performance after finetuning them on the task of action recognition. A more informative task should serve as a better task for pre-training.
\par \noindent \textbf{Two Close}: In this task two frames $(f_b,f_d)$ (with high motion) are considered to be temporally close if $|b - d| < \tau$ for a fixed temporal window $\tau = 30$.
\par \noindent \textbf{Two Order}: Two frames $(f_b,f_d)$ are considered to be correct if $b < d$. Otherwise they are considered incorrect. $|b-d| < 30$.
\par \noindent \textbf{Three Order}: This is the original temporal ordering task we proposed in Section~\ref{sec:intuition}. We consider the 3-tuple $(f_b,f_c,f_d)$ to be correct only if the frames obey either ordering $b < c < d$ or $b > c > d$.

We also compare against standard baselines for unsupervised learning from video.
\par \noindent \textbf{DrLim}~\cite{hadsell2006dimensionality}: As Equation~\ref{eqn:drlim} shows, this method enforces temporal smoothness over the learned features by minimizing the $l_{2}$ distance $d$ between representations (\texttt{fc7}) of nearby frames $f_b,f_d$ (positive class or $c=1$), while requiring frames that are not close (negative class or $c=0$) to be separated by a margin $\delta$. We use the same samples as in the `Two Close' baseline, and set $\delta=1.0$~\cite{mobahi2009deep}.
\begin{equation}
\label{eqn:drlim}
L(f_b,f_d) =  \mathds{1}(c=1) d(f_b,f_d) + \mathds{1}(c=0) \max(\delta-d(f_b,f_d),0) 
\end{equation}
\par \noindent \textbf{TempCoh}~\cite{mobahi2009deep}: Similar to the DrLim method, temporal coherence learns representations from video by using the $l_{1}$ distance for pairs of frames rather than the $l_{2}$ distance of DrLim.
\par \noindent \textbf{Obj. Patch}~\cite{wang-gupta}: We use their publicly available model which was unsupervised pre-trained on videos of objects. As their patch-mining code is not available, we do not do unsupervised pre-training on UCF101 for their model.

All these methods (except~\cite{wang-gupta}) are pre-trained on training split 1 of UCF101 without action labels, and then finetuned on test split 1 of UCF101 actions and HMDB51 actions. We compare them in Table~\ref{tbl:twoframes}. Scratch performance for test split 1 of UCF101 and HMDB51 is $39.1\%$ and $14.8\%$ respectively. The tuple verification task outperforms other sequential ordering tasks, and the standard baselines by a significant margin. We attribute the low number of~\cite{wang-gupta} to the fact that they focus on object detection on a very different set of videos, and thus do not perform well on action recognition.

\begin{table}[!t]
\centering
\caption{\footnotesize{We compare the unsupervised methods defined in Section~\ref{sec:twoframes} by finetuning on the UCF101 and HMDB51 Action recognition (split 1 for both). Method with * was not pre-trained on action data.}}
\label{tbl:twoframes}
\setlength{\tabcolsep}{0.4em}
\begin{tabular}[b]{@{}lcccccc@{}}
\toprule
\multirow{2}{*}{Unsup Method $\rightarrow$} & Two & Two &  DrLim & TempCoh & Three & Obj. Patch* \\
 & Close & Order &  \cite{hadsell2006dimensionality} & \cite{mobahi2009deep} & Order (Ours) & \cite{wang-gupta}\\
\toprule
Acc. UCF101 & 42.3 & 44.1 & 45.7 & 45.4 & \textbf{50.9} & 40.7\\
Acc. HMDB51 & 15.0 & 16.4 & 16.3 & 15.9 & \textbf{19.8} & 15.6\\
\bottomrule
\end{tabular}
\end{table}
\begin{table}[!ht]
\centering
\caption{\footnotesize{Results of using our unsupervised pre-training to adapt existing image representations trained on ImageNet. We use unsupervised data from training split 1 of UCF101, and show the mean accuracy (3 splits) by finetuning on HMDB51.}}
\label{tbl:imnet2hmdb51}
\begin{tabular}[b]{@{}cccc@{}}
\toprule
Initialization & Mean Accuracy \\
\toprule
Random & 13.3\\
\arrayrulecolor{Gray}
\midrule
\textbf{(Ours)} Tuple verification & 18.1\\
\arrayrulecolor{Gray}
\midrule
UCF sup. & 15.2\\
ImageNet & 28.5 \\
\textbf{(Ours)} ImageNet + Tuple verification & \textbf{29.9}\\
\arrayrulecolor{Gray}
\midrule
ImageNet + UCF sup. & 30.6\\
\arrayrulecolor{black}
\bottomrule
\end{tabular}
\end{table}

\subsection{Combining unsupervised and supervised pre-training}
\label{sec:unsup-sup}
We have thus far seen that unsupervised pre-training gives a significant performance boost over training from random initialization. We now see if our pre-training can help improve existing image representations. Specifically, we initialize our model using the weights from the ImageNet pre-trained model and use it for the tuple-prediction task on UCF101 by finetuning for $10$k iterations. We hypothesize this may add complementary information to the ImageNet representation. To test this, we finetune this model on the HMDB51~\cite{hmdb51} action recognition task. We compare this performance to finetuning on HMDB51 without the tuple-prediction task. Table~\ref{tbl:imnet2hmdb51} shows these results.

Our results show that combining our pre-training with ImageNet helps improve the accuracy of the model (rows 3, 4). Finally, we compare against using multiple sources of supervised data: initialized using the ImageNet weights, finetuned on UCF101 action recognition and then finetuned on HMDB51 (row 5). The accuracy using all sources of supervised data is only slightly better than the performance of our model (rows 4, 5). This demonstrates the effectiveness of our simple yet powerful unsupervised pre-training.

\section{Pose Estimation Experiments}
\label{sec:pose}
The qualitative results from Sec~\ref{sec:qual-res} suggest that our network captures information about human pose. To evaluate this quantitatively, we conduct experiments on the task of pose estimation using keypoint prediction.

\noindent \textbf{Datasets and Metrics:}  We use the FLIC (full)~\cite{modec13} and the MPII~\cite{andriluka14cvpr} datasets. For FLIC, we consider 7 keypoints on the torso: head, $2\times$ (shoulders, elbows, wrists). We compute the keypoint for the head as an average of the keypoints for the eyes and nose. We evaluate the Probability of Correct Keypoints (PCK) measure~\cite{yang2013articulated} for the keypoints. For MPII, we use all the keypoints on the full body and report the PCKh@0.5 metric as is standard for this dataset.


\noindent \textbf{Model training:} We use the CaffeNet architecture to regress to the keypoints. We follow the training procedure in~\cite{deeppose}\footnote{Public re-implementation from \url{https://github.com/mitmul/deeppose}}. For FLIC, we use a train/test split of 17k and 3k images respectively and finetune models for 100k iterations. For MPII, we use a train/test split of 18k and 2k images. We use a batch size of $32$, learning rate of $5\times10^{-4}$ with AdaGrad~\cite{duchi2011adaptive} and minimize the Euclidean loss ($l_2$ distance between ground truth and predicted keypoints). For training from scratch (Random Init.), we use a learning rate of $5\times10^{-4}$ for $1.3$M iterations. 

\noindent \textbf{Methods:} Following the setup in Sec~\ref{sec:unsup-finetune}, we compare against various initializations of the network. We consider two supervised initalizations - from pre-training on ImageNet and UCF101. We consider three unsupervised initializations - our tuple based method, DrLim~\cite{hadsell2006dimensionality} on UCF101, and the method of~\cite{wang-gupta}. We also combine our unsupervised initialization with ImageNet pre-training.

Our results for pose estimation are summarized in Table~\ref{tbl:pose-pck-all}. Our unsupervised pre-training method outperforms the fully supervised UCF network (Sec~\ref{sec:unsup-finetune}) by $+7.6\%$ on FLIC and $+2.1\%$ on MPII. Our method is also competitive with ImageNet pre-training on both these datasets. Our unsupervised pre-training is complementary to ImageNet pre-training, and can improve results after being combined with it. This supports the qualitative results from Sec~\ref{sec:qual-res} that show our method can learn human pose information from unsupervised videos.


\begin{table}[!t]
\centering
\footnotesize{
    \caption{Pose estimation results on the FLIC and MPII datasets.}
    \label{tbl:pose-pck-all}
    \begin{tabular}{@{}cccccccccccc@{}}
    \toprule
    Init. & \multicolumn{7}{c}{PCK for FLIC} && \multicolumn{3}{c}{PCKh@0.5 for MPII}\\
    \arrayrulecolor{Gray}
    \cmidrule(l{-0.1cm}r{2.0pt}){2-8}
    \cmidrule(l{2.5pt}){9-12}
    \arrayrulecolor{black}
     & wri & elb & sho & head && Mean & AUC && Upper & Full & AUC\\
    \midrule
    Random Init. & 53.0 & 75.2 & 86.7 & 91.7 && 74.5 & 36.1 && 76.1 & 72.9 & 34.0\\
    \arrayrulecolor{Gray}
    \midrule
    Tuple Verif. & \textbf{69.6} & \textbf{85.5} & \textbf{92.8} & \textbf{97.4} && \textbf{84.7} & \textbf{49.6} && \underline{\textbf{87.7}} & \textbf{85.8} & \textbf{47.6}\\
    Obj. Patch\cite{wang-gupta} & 58.2 & 77.8 & 88.4 & 94.8 && 77.1 & 42.1  && 84.3 & 82.8 & 43.8\\
    DrLim\cite{hadsell2006dimensionality} & 37.8 & 68.4 & 80.4 & 83.4 && 65.2 & 27.9 && 84.3 & 81.5 & 41.5\\
    \midrule
    UCF Sup. & 61.0 & 78.8 & 89.1 & 93.8 && 78.8 & 42.0 && 86.9 & 84.6 & 45.5\\
    ImageNet & 69.6 & 86.7 & 93.6 & 97.9 && 85.8 & 51.3 && 85.1 & 83.5 & 47.2\\
    \midrule
    ImageNet + Tuple & \underline{\textbf{69.7}} & \underline{\textbf{87.1}} & \underline{\textbf{93.8}} & \underline{\textbf{98.1}} && \underline{\textbf{86.2}} & \underline{\textbf{52.5}} && \textbf{87.6} & \underline{\textbf{86.0}} & \underline{\textbf{49.5}}\\
    \arrayrulecolor{black}
    \bottomrule
    \end{tabular}
}
\end{table}


\section{Discussion}
In this paper, we studied unsupervised learning from the raw spatiotemporal signal in videos. Our proposed method outperforms other existing unsupervised methods and is competitive with supervised methods. A next step to our work is to explore different types of videos and use other `free' signals such as optical flow. Another direction is to use a combination of CNNs and RNNs, and to extend our tuple verification task to much longer sequences. We believe combining this with semi-supervised methods~\cite{MisraSSL15,liang2015towards} is a promising future direction.

{\footnotesize
\noindent \textbf{Acknowledgments:} The authors thank Pushmeet Kohli, Ross Girshick, Abhinav Shrivastava and Saurabh Gupta for helpful discussions. Ed Walter for his timely help with the systems. This work was supported in part by ONR MURI N000141612007 and the US Army Research Laboratory (ARL) under the CTA program (Agreement W911NF-10-2-0016). We gratefully acknowledge the hardware donation by NVIDIA.
}

\clearpage

\bibliographystyle{splncs}
\bibliography{myRefs}
\clearpage
\section{Appendix}
We present extra results and analysis of our method.

\subsection{More qualitative results}
\par
\begin{figure}[!b]
\centering
\includegraphics[width=0.9\textwidth]{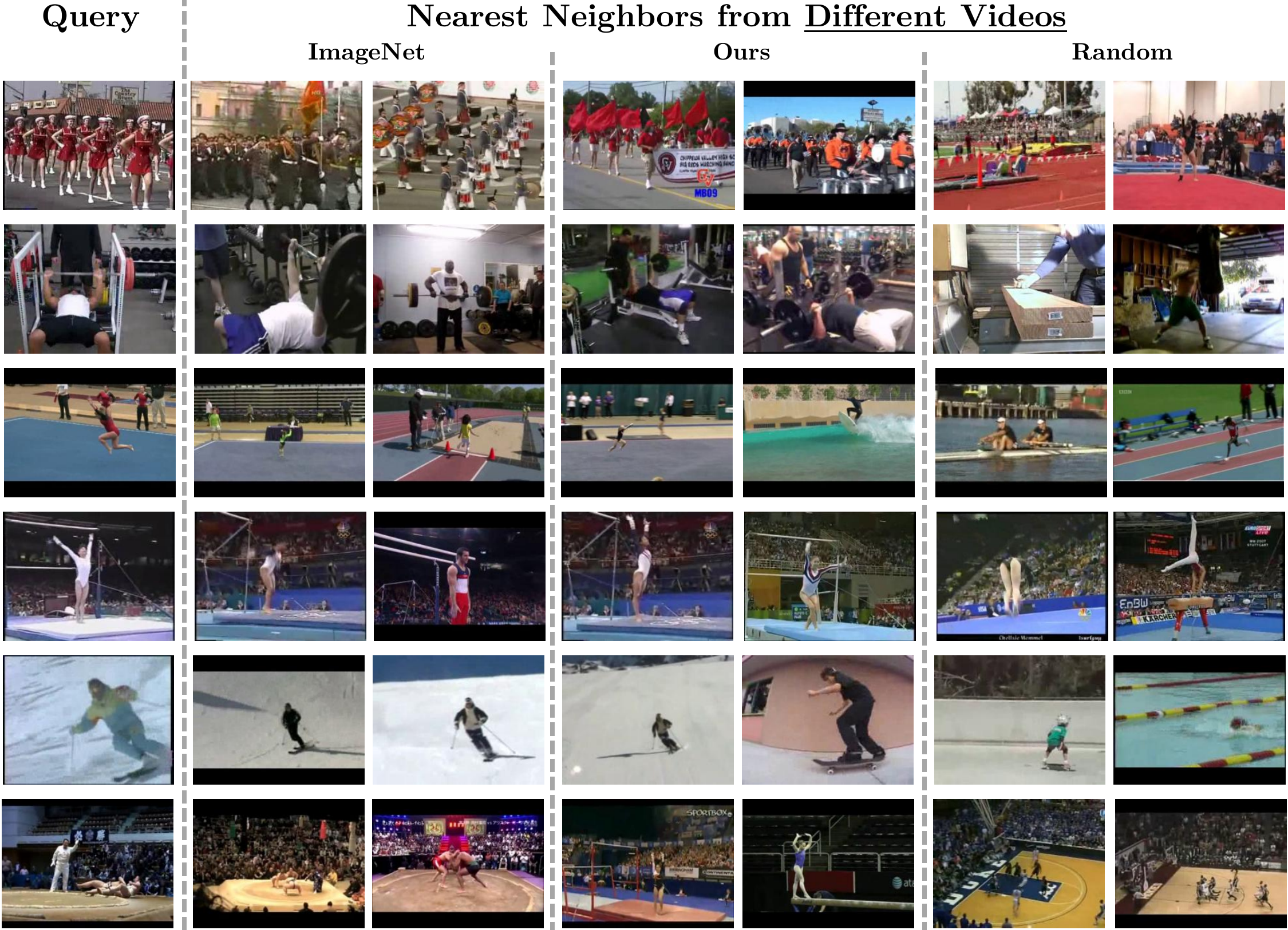}
\caption{\footnotesize{We compute nearest neighbors using \texttt{fc7} features on the UCF101 dataset. We compare these results across three networks: pre-trained on ImageNet, pre-trained on our unsupervised task and a randomly initialized network. We choose a input query frame from a clip and retrieve results from other clips in the dataset. Since the dataset contains multiple clips from the same video we get near duplicate retrievals. We remove these duplicates, and display results. While ImageNet focuses on the high level semantics, our network captures the human pose.}}
\label{fig:nn-sup}
\end{figure}

 \noindent \textbf{Nearest Neighbors:} We provide more visualizations of nearest neighbors in Figure~\ref{fig:nn-sup}, similar to Section 4.3 of the main paper.

\par
\begin{figure}[!b]
\centering
\includegraphics[width=0.9\textwidth]{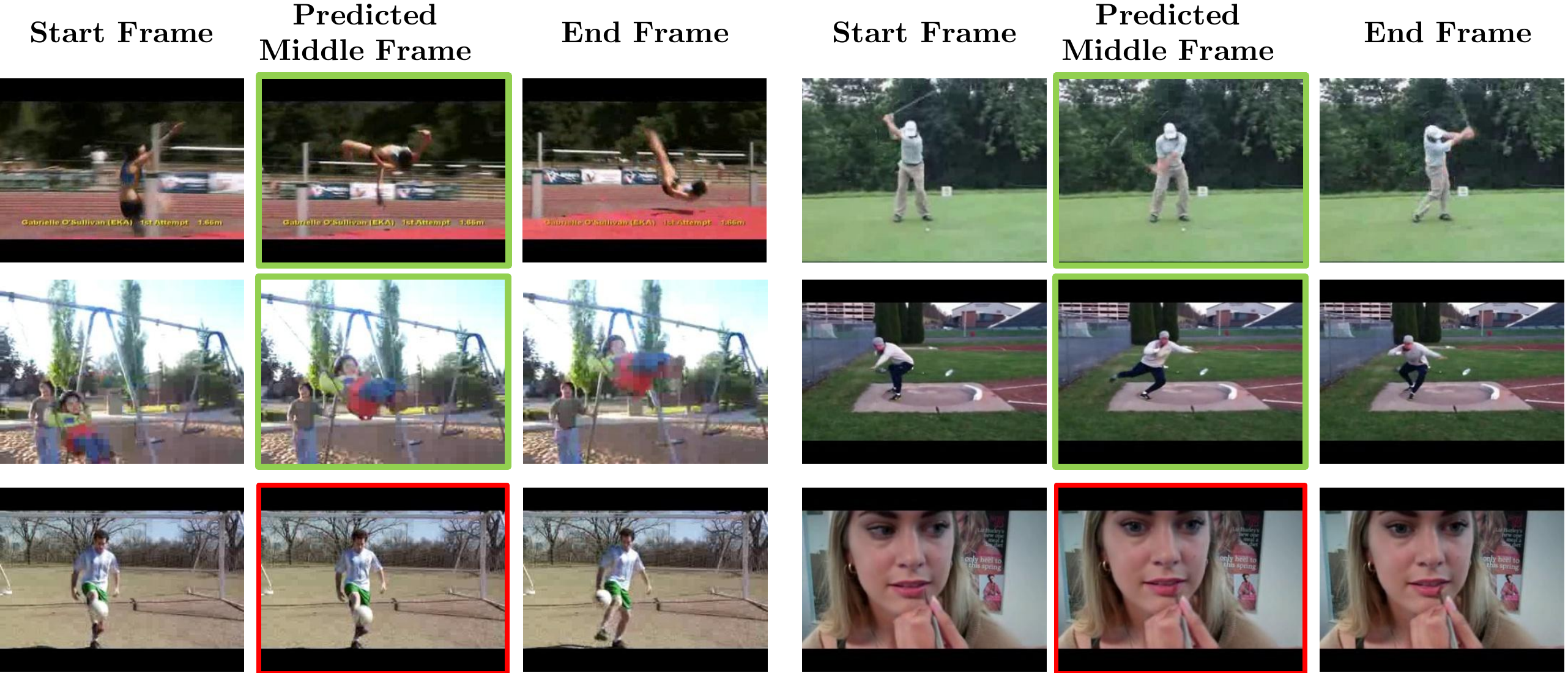}
\caption{\footnotesize{Given a start and an end frame on a held out video, we use our network to find a middle frame. The first two rows demonstrate correct predictions, with the network correctly predicting motion for cyclical cases, \eg, a child on a swing. The last row depicts wrong predictions.}}
\label{fig:fill-blanks}
\end{figure}

 \noindent \textbf{Fill in the blanks:} Given a start and an end frame, we use our unsupervised network to find a middle frame from the input video. We compute such results on held out videos and show them in Figure~\ref{fig:fill-blanks}. Our network is able to correctly predict frames that should temporally lie between a given start and end frame. For cyclical actions with large motion, \eg, a child on a swing (row 2), our network resolves directional ambiguity (is the swing going up or down). The last row shows failure cases which lack motion (applying makeup) or have only small moving objects (soccer ball).

\subsection{Control experiments}
In these experiments we control for certain variations like batch normalization, number of iterations etc. in our method and the baseline methods. These experiments will help us tease apart gains over the baselines that are due to these variations vs. gains due to our unsupervised pre-training.

\par \noindent \textbf{Control for more iterations:}
Our unsupervised training method is trained for a larger number of iterations (without action labels). This gives it an advantage over the baselines that are trained from scratch or finetuned from other datasets (like ImageNet~\cite{ImageNet,ILSVRC15}). To control for this, we run the baseline methods, both scratch and finetuning, for a higher number of iterations (200k vs. 80k for scratch, 40k vs. 20k for finetune) than in~\cite{simonyan-2stream}, and report the highest accuracy.

\par \noindent \textbf{Control for batch normalization:}
We use batch normalization~\cite{batchnorm} for training our triple-Siamese network. Since the other baseline methods from~\cite{simonyan-2stream} predate the batch normalization method, we first re-trained the baseline models using batch normalization. For lack of space, these numbers are reported in the supplementary material.
As also reported by~\cite{doersch-context}, we observed that using batch normalization consistently gave about $1\%$ worse accuracy. Thus, we report baseline numbers without batch normalization.

\subsection{Pose Estimation}
\begin{figure}[!b]
\centering
	\captionof{figure}{\footnotesize{PCK values for the wrist and elbow keypoints on the FLIC dataset}}
	\label{fig:pck}
	\includegraphics[width=0.45\textwidth]{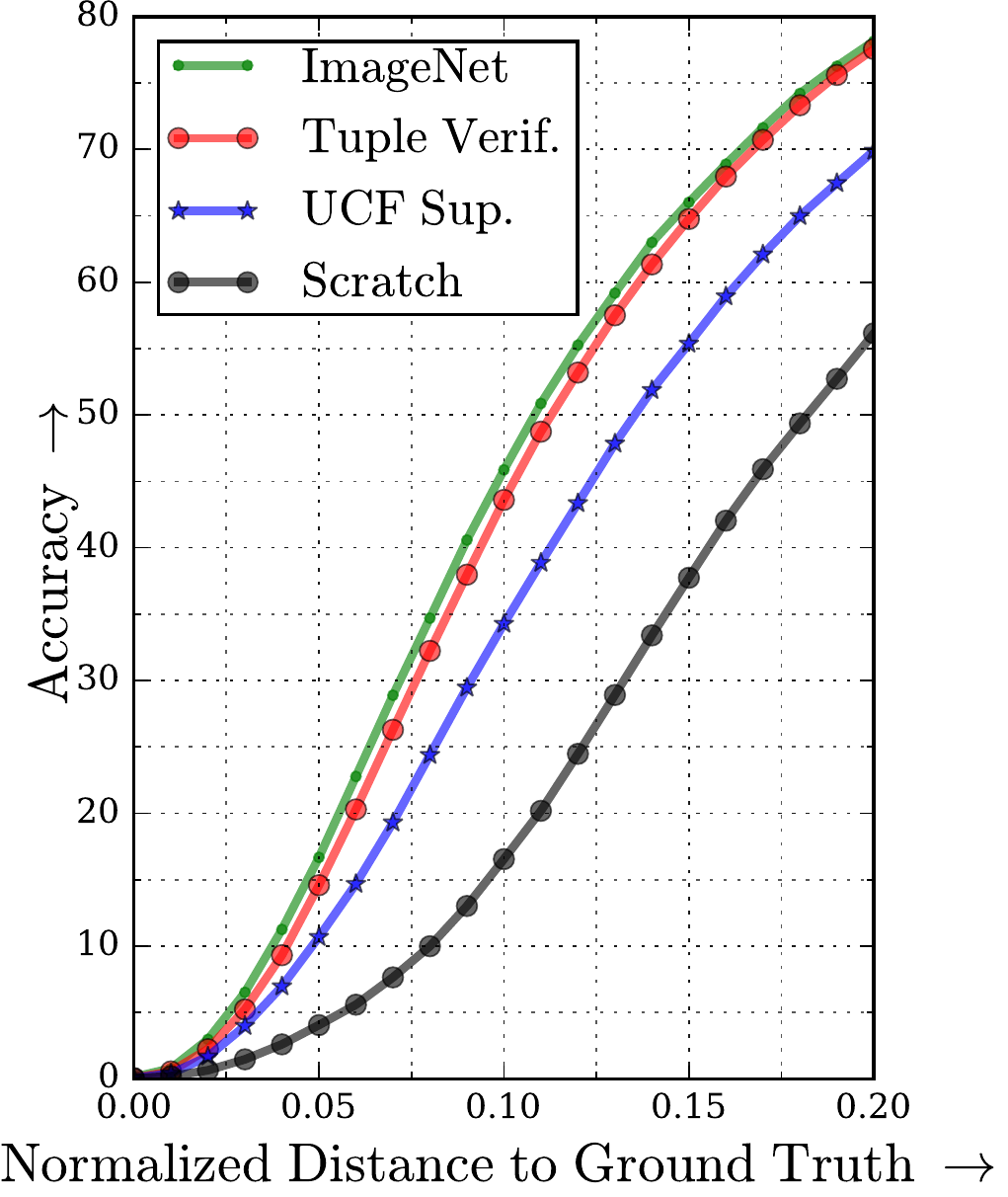}
\end{figure}

We evaluate the Probability of Correct Keypoints (PCK) measure~\cite{yang2013articulated} on the wrist and elbow keypoints for the FLIC-full dataset, as used in~\cite{pfister2015flowing} (Figure~\ref{fig:pck}). PCK measures the correctness of predicted keypoints by varying the distance threshold at which they are considered correct.

\end{document}